# L-System Genetic Encoding for Scalable Neural Network Evolution: A Comparison with Direct Matrix Encoding

Alexander Stuy
M.S. Computer Science 1994, Florida State University
astuy@bio.fsu.edu

Nodin Weddington
B.S. Physics 2002, Florida State University
nweddington@fsu.edu

## Abstract

An artificial world of barriers and plains scattered with food is used to test the feasibility of using genetic algorithms to optimize hebbian neural networks to perform on problems without a priori knowledge of the problem domain. A formal L-System based genetic alphabet for neural networks, titled Lsys, and a neural network genetic modeling tool titled Wp1hgn are introduced. Lsys and Matrix neural network topology genetic encoding methods are compared across 24 experimental runs. Lsys encoding achieved a mean maximum food count of 3802 ± 197 at generation 1000 across 8 runs with varied parameters, compared to 1388 ± 610 for Matrix encoding, a 2.74x performance advantage with an 8.5-fold improvement in consistency as measured by coefficient of variation (5.2% vs 44.0%). All 8 Lsys populations successfully learned to navigate the environment, while 4 of 8 Matrix populations failed to achieve competitive performance at any point during 1000 generations. When transferred to a novel maze environment, Lsys populations demonstrated immediate robust generalization, achieving a mean maximum food count of 2455 ± 176 compared to 422 ± 212 for Matrix populations, a 5.82x advantage that exceeded the training world performance gap. A MatrixLSG control condition, in which initial populations were generated using Lsys genotypes and then evolved using Matrix operators, demonstrated that the performance advantage of Lsys encoding derives primarily from the genetic algorithm operating on the compressed symbolic Lsys alphabet throughout evolution rather than from initial population structure. Lsys encoding is shown to provide faster convergence, higher peak performance, dramatically greater reliability, and superior generalization to novel environments compared to Matrix encoding across all experimental conditions tested.

## 1 Introduction

Artificial neural networks and genetic algorithms are both inspired by the success of their biological counterparts. Speed increases in computers in recent years have allowed scientists to combine the software models of these processes. Genetic algorithms have proved capable of optimizing large complex structures, and a artificial neural network is just such a structure. Neural network optimization methods manipulate the number of neurons and the connections between them (network topology) in order to generate a network that learns better than one that is unoptimized. By encoding the network topology into a genotype, genetic algorithms can be used to perform the optimization.

Neural networks can be genetically optimized to perform on traditional neural network problems such as pattern recognition and input/output pattern learning. Even more interesting however is that networks can be evolved to solve non-continuous functions without any apriori knowledge of the function. In their 1988 paper Stefano Nolfi and Domenico Parisi describe genetically optimized back-propagation neural networks which demonstrate an ability to learn to collect food in an artificial environment [12]. A normal back propagation network is combined with a genetic neural network which supplies the teaching input to the back propagation network. Biological neurons have been shown to employ a form of learning known as hebbian learning [7]. Artificial hebbian networks can be used for feature identifying, data clustering and object recognition [6].

### Overview of Experiment

The first purpose of the experiment that is the basis of this paper is an attempt to optimize a hebbian neural network for existence in an artificial environment. Possible applications of the technique include robot control and system control, however more traditional neural network problems could also be programmed.

Performance of the artificial hebbian networks will be observed by introducing each network into an artificial environment that the network must navigate while searching for food. Network performance is measured as the amount of food found by the network. If the genetic algorithm is successful in optimizing the hebbian neural networks the average performance of the networks will increase through successive generations.

Further the experiment will measure the performance of two different methods for encoding the network topologies. By keeping all factors constant except for the encoding method the experiment will determine whether the encoding method used to encode the neural network topology affects the performance of the genetic algorithm, and if so, which type of encoding methods offer the best performance. Encoding method one, hereafter referred to as Matrix encoding, is standard matrix encoding of a neural network, connectivity and weight vectors. Encoding method two, hereafter referred to as Lsys encoding, is based on L-Systems, which have been shown to generate patterns resembling patterns produced by growth in biological organisms [10] and is used to model the growth of biological neuron structures [17].

# 2 Background

## 2.1 Neural Networks

Artificial neural networks were inspired by the computing ability of biological neural networks. The human brain performs feats, such as seemingly parallel classification of multiple objects, in comparable time to today's fastest computers. The feat is remarkable as the speed at which biological neurons operate is seven orders of magnitude slower than computer semiconductor gates [4]. The brain performs such feats by operating ~$10^{11}$ neurons of at least 14 distinct types in parallel [4]. Artificial neural networks consist of computational neuron simulations, also with multiple possible types of neurons.

### 2.1.1 Biological Neural Networks

A biological neural network (BNN) is composed of specialized cells called *neurons*. The main components of biological neurons are a *cell body*, *nucleus*, *axons* and *dendrites* (Figure 1). The dendrites receive signals from the nervous system or from the axons of other neurons. Depending on the input from its dendrites the neuron generates an impulse known as an action potential [4]. This impulse propagates as a wave down the axon, providing a signal to the dendrites of other neurons connected to the axon's branches. Learning in the brain can occur by strengthening or weakening the connections between axons and dendrites. In 1949 Donald Hebb postulated a learning rule, dubbed hebbian learning , as a model of learning for biological neurons [7]. The hebbian model strengthens the weight on a connection whenever the neurons on both sides of the connection fire together.

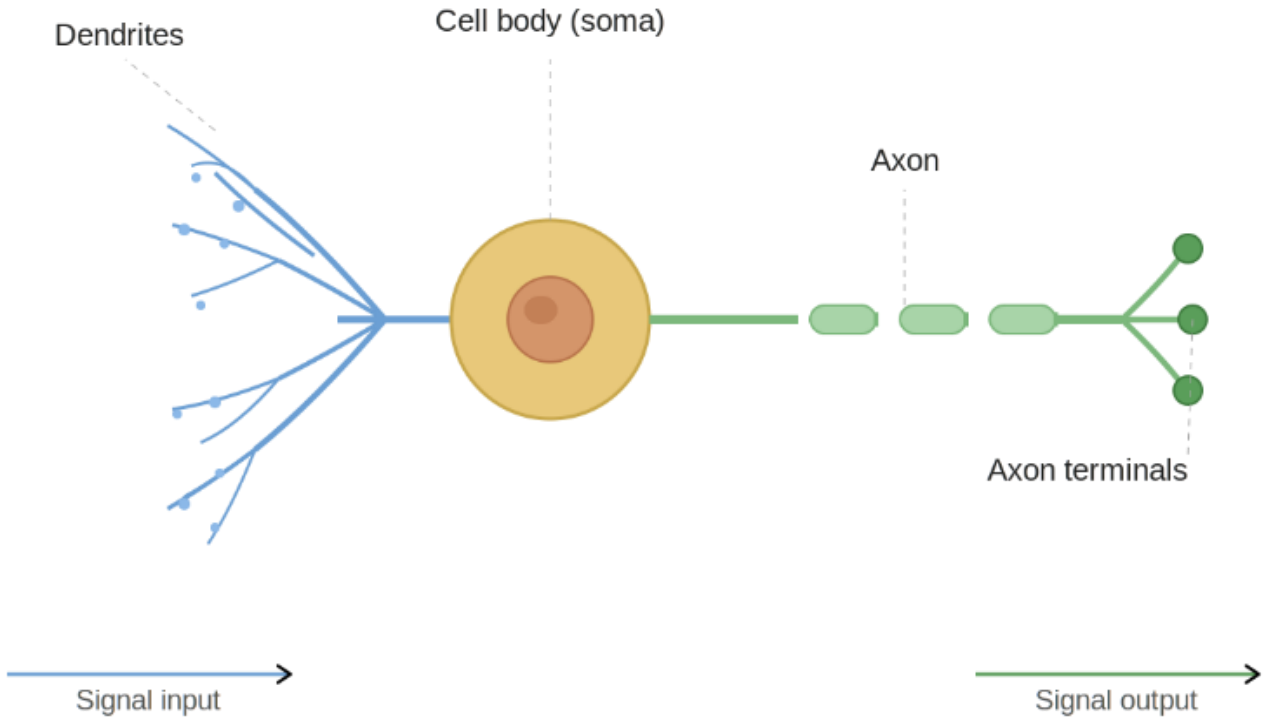

Figure 1: Biological Neuron.

### 2.1.2 Artificial Neural Networks

Like a BNN an artificial neural network (ANN) consists of one or more interconnected neurons. The neurons in an ANN can be described mathematically :

E : input vector to neuron.
w : weight vector for scaling input vector to neuron.
f : firing function, functions used include threshold, linear and sigmoidal functions.
V : output of neuron.

An artificial neuron with i inputs (dendrites) is fired with the formula :

$$V = f(\sum_{k=1..i}(E_k * w_k))$$

One of the first models of an artificial neuron was the McCulloch/Pitts neuron. In the McCulloch/Pitts model the firing function f is a threshold binary function. Each neuron outputs a one if V is above a certain threshold, else the output is zero [6]. Although this seems simple, McCulloch and Pitts proved that a synchronous assembly of these neurons with the proper weight vector w can realize any computable function.

In order for ANN's to learn the values on the weight vector w are adjusted. The value of the adjustment for each $w_i$ is termed $\Delta w_i$. ANN's can be put into two categories. Those that use supervised learning to obtain $\Delta w$ and those that use unsupervised learning to obtain $\Delta w$. With supervised learning the network is supplied a set of training inputs, along with the correct output for each input. The network uses the input output pairs to adjust w to minimize network error on the training set. This performance carries over to inputs not in the training set by virtue of a neural network's ability to generalize. Unsupervised learning ANN's are only supplied with input patterns. The network clusters or classifies the input data into categories. Therefore there must be meaningful redundancies in the input data. Unsupervised networks can perform functions such as clustering of data, encoding and feature mapping, with the architecture of the network regulating the function the network performs [6].

One of the most widely used types of neural networks is the backpropagation network. The output of neurons in a backpropagation network is analog rather than digital, and the firing function f is typically a sigmoidal function. The name backpropagation describes the (supervised) process of propagating the error between expected and actual output backward through the net in order to obtain $\Delta w$'s for each neuron.

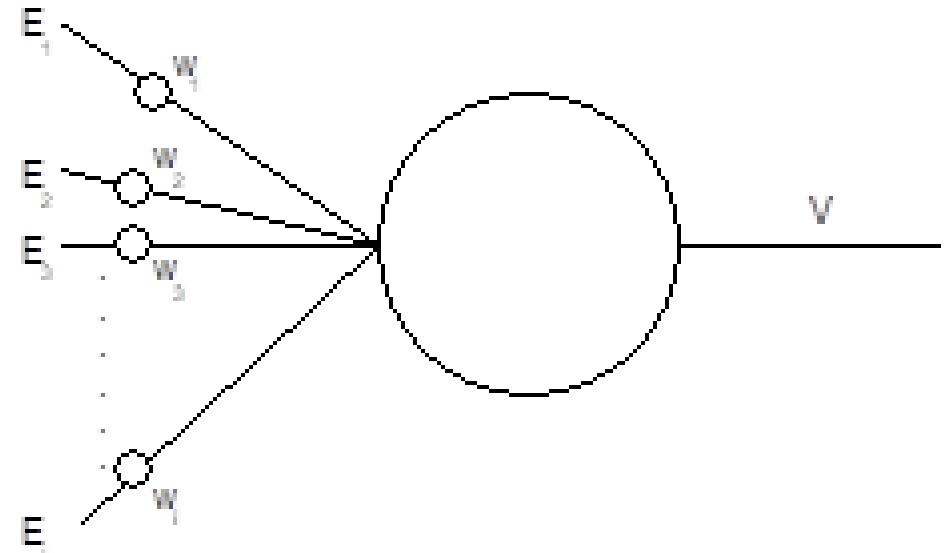

Figure 2 : artificial neuron, $V = f(\sum_{k=1..i}(E_k * w_k))$.

**Hebbian Neural Networks**

In 1949 Donald Hebb postulated a learning rule, dubbed hebbian learning , as a model of learning for biological neurons [7]. The hebbian model strengthens the weight on a connection whenever the neurons on both sides of the connection fire together. Using the notation introduced at the beginning of 2.2.1 the model can be expressed as :

$$\Delta w_i = nVE_i$$

Hebbian neural networks fall into the unsupervised learning category. The information required to update $\Delta w$ is the input to the neuron and the output of the neuron. After sufficient update of the connection weights the output V of each neuron becomes a scalar measure of familiarity of the input E. Output V is directly related to the probability of receiving input E. Hebbian networks have been applied as feature detectors, being given data much as would be received from a retina [6]. Layered hebbian networks have been proposed as a model of self-organization in the visual cortex. Variations of the original hebbian update rule include Oja's rule :

$$\Delta w_i = nV(E_i - Vw_i),$$

Oja's rule avoids unlimited growth in the weights on connections which occurs under Hebb's original rule[6].

## 2.2 Genetic Algorithms

Genetic algorithms (GAs) are computer optimization methods inspired by molecular genetics and Darwin's theory of natural selection. All GAs, both biological and artificial, are search algorithms. In nature natural selection favors organisms with the greatest propagative ability. An organisms propagative ability is a function of the organism's genotype (DNA) and the organism's environment. Biological genetic algorithms search the domain space of possible DNA sequences for ever more fit sequences, in relation to the environment in which the DNA finds itself expressed. Ability to propagate in an environment can be equated to a fitness function for the organisms which inhabit the environment.

GAs operate on genotypes, which are composed of a set of symbols. Most life on earth uses the bases guanine, cytosine, adenine and thymine, a four symbol alphabet commonly abbreviated to G, C, A and T. Artificial GAs commonly use binary encoding methods, although larger alphabets have been used.

A simple artificial GA is composed of the operators : selection, crossover, and mutation.

**selection**

The initial generation of genotypes is generated randomly. After the initial generation is generated the selection operator chooses from the current population the genotypes that will reproduce to form each succeeding generation. A fitness function is used to measure the fitness of each genotype. The parameters to the function are decoded from the genotype. The function used is based on the problem that one is attempting to solve with the genetic algorithm. Many different procedures can be used to perform selection, all involve choosing, either stochastically or deterministically, the fittest genotypes from the population. A commonly used selection procedure is roulette wheel selection. After the fitness of all genotypes has been determined a roulette wheel is generated with each genotype receiving a slot sized in proportion to the genotype's fitness in relation to the average fitness of the entire population. Selection into the reproduction genepool is then determined by simple spins of the roulette wheel.

**crossover**

The crossover operator combines two genotypes chosen for reproduction by splitting each genotype into two or more pieces and combining pieces from both genotypes into a new genotype. Typically not all genotypes selected for reproduction are subject to crossover. A parameter termed the probability of crossover is used to stochastically determine which genotypes are candidates.

**mutation**

Following selection and crossover genotypes are typically subjected to a mutation operator. The mutation operator scans each gene in the genotype and with a probability termed the probability of mutation, flips the gene to another symbol in the genetic alphabet. Mutation helps to maintain population diversity

**Pseudo code for a simple GA**

*Generate the first generation of genotypes randomly*
*Loop*
*Measure fitness of each genotype.*
*Select genotypes for reproduction based on fitness.*
*Construct next generation of genotypes from selected genotypes using the crossover and mutation operators.*
*Until stop-condition*

The process loops until a genotype performs to an acceptable level of fitness, or until the increase of fitness in successive generations grinds to a halt. The second condition can occur due to a number of causes. The two most common are premature convergence and mutation stall.

**premature convergence**

After the initial random generation of the population in a GA it is expected that most genotypes will prove to have a very poor fitness, as they are a random attempt at a solution. However often a few of genotypes will have fitness far above the rest of the population, and although the fitness these genotypes is not the global maximum achievable they will nevertheless dominate the population due to their relative fitness. The GA then becomes trapped at or near this fitness level due to a loss of all other genes from the population (premature convergence).

**mutation stall**

The mutation operator scans each gene in a genotype and with a preset probability flips the gene to another character in the genetic alphabet. While mutation increases genetic diversity in the population mutation also carries the risk of mutating a beneficial gene, resulting in a decrease in fitness. If the mutation rate is set too high the destructive affect can cause the GA's performance to level off after a sufficient level of complexity has arisen in the genotypes (mutation stall).

### 2.2.1 Alphabets

In order to use genetic algorithms to solve a given problem a method for encoding the parameters of the problem (fitness function) into a genotype must be chosen. The smaller the cardinality of the genetic alphabet chosen the more schemata the resulting population will contain, although schemata which cross over parameter encodings are of questionable use. The principle of minimal alphabets states : **The user should select the smallest alphabet that permits a natural expression of the problem** [8]. Parameters are often coded into a binary representation and the resulting binary strings are combined to form a genotype. However non binary encoding have been successfully employed in genetic algorithms, including genetic algorithms used to optimize neural networks [5].

Genetic algorithms have been applied to a wide variety of problems such as pattern recognition, classifier systems and structural optimization. A.K. Dewdey used genetic algorithms to evolve finite automata, dubbed flibs, which predict binary number sequences [2]. Classifier systems also respond well to genetic manipulation. Evolved classifiers systems have been able to perform object recognition in noisy environments [13]. Simulated robots which use genetically selected classifiers to produce output have been evolved to chase a simulated light source.[3]

## 2.3 L-Systems

Lindenmayer systems or L-Systems were introduced by biologist Aristid Lindenmayer in 1968 as a method for describing the morphology of growth in biological systems, particularly plants. L-Systems are a type of context free grammar studied in formal language theory. L-Systems have been widely used to model plant growth and to generate computer graphics. One of the first attempts of modeling biological systems with L-Systems was performed by D. Fritjers and A. Lindenmayer, who modeled the growth and flowering of the plant *Aster novae-angliae* with a L-System [15]. L-Systems can also be used to encode neural network topologies, generating connectivity patterns in the networks similar to connectivity patterns seen in regions of the brain [10].

L-Systems consist of a finite *alphabet* of symbols and a set of rules or *productions*, each of which maps one symbol or string of symbols from the alphabet to another symbol or string of symbols. The alphabet usually consists of upper and lower case letters, though other alphabets can be used. The set of characters {S, A, B, C, D, a, b, c, d, e, f} is a valid L-System alphabet. A valid production using the above defined alphabet is : S -> BCdA. All L-Systems must have a start symbol or an identity production. S is commonly used to denote the start symbol and is used only on the left side of the identity production. Symbols that are never found on the left hand side of a production are termed terminals, as they represent an end to the production process. Non terminals appear on the left hand side of at least one production. Any string of symbols produced by the execution of the productions of a L-System is termed a *word* of the L-System. The set of all possible words produced by an L-System is termed the *language* of the L-System.

Figure 3 shows an example L-System as well as a graph representation of the L-System. This particular L-System models the growth of multi-cellular filament found in the blue-green bacteria *Anabaena catenula* [14]. The symbols a and b correspond to individual cells, describing the cells propensity to divide, while the subscripts l and r specify the position of the newborn cell after division.

S : $a_r$

P1 : $a_r$ -> $a_l b_r$

P2 : $a_l$ --> $b_l a_r$

P3 : $b_r$ -> $a_r$

P4 : $b_l$ -> $a_l$

Sequence of words generated :

$a_r$
$a_l b_r$
$b_l a_r a_r$
$a_l a_l b_r a_l b_r$
$b_l a_r b_l a_r a_r b_l a_r a_r$
....

Figure 3: Simulated growth of filament from bluegreen algae *Anabaena catenula*

L-Systems can also used to describe biological neural structures. Ascoli and Krichmar introduced L-Neuron, a software tool that uses L-System based recursive rules to generate anatomically accurate models of dendritic morphology [16]. L-Neuron demonstrates that the branching structure of biological neurons, including Purkinje cells, pyramidal cells, and motoneurons, can be described parsimoniously using L-System productions, achieving significant data compression while preserving the essential structural properties of each morphological class. The authors note that thousands of neurons from a single morphological class, requiring megabytes of storage in explicit Cartesian format, can be described completely with a small set of L-System parameters, a compression principle directly analogous to the $O(\log_2 n)$ scaling property of the Lsys encoding method introduced in this paper. The convergence of two independent lines of research, L-System based genetic encoding of network topology and L-System based description of dendritic morphology, suggests that the hierarchical recursive structure of L-Systems captures something fundamental about how biological neural connectivity is organized. This biological plausibility may help explain why Lsys encoded networks exhibit more organic emergent behaviors than matrix encoded networks, including the development of temporally extended responses and adaptive navigation strategies that parallel behaviors observed in biological organisms.

## 2.4 Genetic Algorithms and Neural Networks

The result of combining genetic algorithms with neural networks is termed a genetic neural network (GNN). Genetic neural networks consist of four components: genes that specify the network topology, a procedure for constructing the network from the genotype, a fitness function to measure the performance of the network, and genetic operators for creating the new generation of genotypes from the old generation [11]. Genetic algorithms can be used to optimize the number of neurons, the topology of the network, and/or to set the weights on some subset of the networks connections.

Genetic neural networks can learn on two levels. On the individual level each network learns by the weight update rule used (neural network learning). On a population level the networks learn by modifying their topologies over time to decrease the average error in network output of the population. These two processes are known as phenotype and genotype learning, respectively. Genotype learning has been demonstrated to progress at a faster rate when the phenotypes have an ability to learn [9]. This should not be confused with lamarkian evolutionary theory. Only the genotypes are used in the genetic algorithm, what a phenotype learns is not passed on.

In order to use genetic algorithms with neural networks some method of encoding the network topology into a form usable as a genotype must be chosen. Neural networks are often denoted as a binary matrix with size of the number of neurons in the network. A zero in the matrix denotes the lack of a connection between the corresponding neurons, a one denotes a connection. By adding a real component to the matrix the weight information of each connection can also be stored.

Genetic algorithms have demonstrated success both when used to entirely determine network architecture, including weights [5], as well as when combined with learning rules[9].

### 2.4.1 Efficiency of coding methods

Encoding the neural network topology as a matrix for genetic algorithms remains the standard approach in neuroevolution, with recent work continuing to apply matrix-based crossover operators to fully connected layer weight matrices [16]. However some experiments have shown that direct encoding methods may not scale up well compared to more symbolic representations. The size of the matrix based genotypes generally scales very poorly and the number of generations required to achieve maximum performance increases exponentially as the number of neurons encoded by the genotype increases [10]. Graph L-Systems, which generate graphs by rules rather than explicit encodings have been shown to have superior scaling ability, as well as providing a speed up in convergence of the genetic algorithm [10].

The efficiency of an encoding method is often measured in O (Big O) notation. With linear encoding methods such as integer pairs, the genotypes grow at O(m), m = number of connections. Encoding methods based on connectivity matrixes are $O(n^2)$, n = number of neurons, since the number of cells in the matrix is $n^2$. L-System based encoding methods can achieve growth rates in genotypes lower than O(n). The following is a proof that L-System encoding methods can achieve genotype growth as low as $O(\log_2 n)$. The proof uses a simplified version of the Lsys encoding method.

## Proof of $O(\log_2 n)$ growth of L-System genotypes for neural networks

For a given L-System, *l,* three terms related to L-Systems are introduced :

S(*l*) : the number of symbols in the rules of the L-System.

T(*l*) : the number of symbols, (terminals), in the expansion of the L-System productions.

N(*l*) : the maximum number of neurons in the neural network that can be encoded by the L-System.

At this point it should be noted that the standard encoding for a neural network of x neurons uses $x^2$ symbols, (a standard connectivity matrix of ones and zeros). Thus a L-System that expands to T terminal symbols, where the terminal symbols are 0 and 1, encodes a network with sqrt(T) neurons, meaning that N(*l)* = sqrt(T(*l*)).

To prove that L-System encoding methods for neural networks can be $O(\log_2 N(l))$ a series of L-Systems, labeled LS[i], i = 1.. ∞, will be introduced such that :

a. S(LS[i + 1]) = S(LS[i]) + C, for some constant C.

b. T(LS[i + 1]) = 4T(LS[i]).

**Proof**

If

S(LS[i + 1]) = S(LS[i]) + C
T(LS[i + 1]) = 4T(LS[i])
N(LS[i]) = sqrt(T(LS[i]))

then

N(LS[i+1]) = sqrt(T(LS[i + 1]))
= sqrt(4T(LS[i]))

$= 2N(LS[i])$

$S(LS[i]) = S(LS[0]) + i*C$

$N(LS[i]) = 2^{i}N(LS[0])$

Therefore given an L-System, l, which meets conditions a and b above, growth of S($l$) in relation to N($l$) is O($\text{Log}_2 N(l)$).

**The L-System**

In addition to the notation already presented, notation for the L-System in the proof will use the following terms :

t : terminal symbol variable, with possible values of one or zero.

Li : nonterminal symbol of level i, if Li has a subscript it refers to a specific L-System symbol of level i, else it is a symbol variable with possible values being the symbols occuring on the left side of the productions in level i.

The L-Systems LS[i] are defined as follows :

Each L-System LS[i] has i levels. A level consists of a set of Symbols and productions. The productions in each level map the symbols of that level to a set of four symbols in the next level, the last level maps its symbols to the terminal symbols (0, 1).

LS[1] : $L1_1$ -> t t t t

LS[2] : $L1_1$ -> L2 L2 L2 L2

$L2_1$ -> t t t t, $L2_2$ -> t t t t, $L2_3$ -> t t t t, $L2_4$ -> t t t t

LS[3] : $L1_1$ -> L2 L2 L2 L2

$L2_1$ -> L3 L3 L3 L3, $L2_2$ -> L3 L3 L3 L3,

$L2_3$ -> L3 L3 L3 L3, $L2_4$ -> L3 L3 L3 L3

$L3_1$ -> t t t t, $L3_2$ -> t t t t, $L3_3$ -> t t t t, $L3_4$ -> t t t t

.

.

LS[i] : $L1_1$ -> L2 L2 L2 L2

$L2_1$ -> L3 L3 L3 L3, $L2_2$ -> L3 L3 L3 L3,

$L2_3$ -> L3 L3 L3 L3, $L2_4$ -> L3 L3 L3 L3

.

.

$Li_1$ -> t t t t, $Li_2$ -> t t t t, $Li_3$ -> t t t t, $Li_4$ -> t t t t

Statistics for these L-Systems.

LS[1] : S(LS[1]) = 5.
T(LS[1]) = 4.
N(LS[1]) = 2.

LS[2] : S(LS[2]) = 5 + 20.
T(LS[2]) = 16.
N(LS[2]) = 4.

LS[3] : S(LS[3]) = 5 + 20 + 20.
T(LS[3]) = 64.
N(LS[3]) = 8.

LS[4] : S(LS[4]) = 5 + 20 + 20 + 20.
T(LS[4]) = 256.
N(LS[4]) = 16.

The value of S(LS[1]) is five. Each additional level of symbols consists of four rules, 20 symbols. Thus S(LS[i + 1]) = S(LS[1]) + i * 20. 20 is a constant, condition a has been met.

Each nonterminal symbol is expanded to exactly four symbols from the following level therefore T(LS[i + 1]) = T(LS[i]) * 4, condition b has been met.

# 3 The Experiment

**Overview**

To perform the experiment a population of animats is constructed. Each animat consists of a hebbian/genetic ANN, input sensors, and an output register. Based on the contents of the output register the animat has the ability to turn, move and collect food. The animats are introduced, one at a time, into an artificial environment, and allowed to exist for a predetermined life span or until starving (not collecting food for a predetermined amount of time). The environment consists of open terrain with scattered barriers. Food is distributed randomly in the environment before each animat is introduced. Animat performance is measured by the amount of food collected in the animat's life span. The ability to avoid barriers is important to an animat's ability to collect food. Animats that become stuck on a barrier or continuously bump into barriers do not have the opportunity to collect food.

At the end of a generation (all animats have lived in the environment) GAs will be used to produce the next generation of animats from the current generation. The fitness function used for the GA is the performance of the animat in the artificial environment and is measured by the amount of food collected.

The purpose of this experiment is twofold. The first is to determine whether genetic algorithms can optimize hebbian neural network topologies for a given problem. Optimization will be deemed successful if there is significant improvement in network performance from the first generation. The second purpose is to experiment with Matrix vs Lsys encoding to determine whether the genetic encoding method has an effect on performance. The Matrix and Lsys encoding methods will be used on the same fitness function, with all parameters, except those related to the encoding, set the same. Thus any difference in performance should be due to the encoding method being used. For a ANN with n neurons Matrix encoding is a $O(n^2)$ encoding method. Lsys encoding method is a $O(\log_2 n)$ encoding method. L-System based encoding methods have been shown to provide superior performance over other encoding methods [10]. The predicted result is that Lsys encoding will outperform Matrix encoding.

## 3.1 Design

### 3.1.1 The environment

An artificial environment is used to test each animat's neural network performance. The environment consists of a 110 by 90 grid of cells. Each cell contains information about its terrain, food content, and occupant. There are two types of terrain, open and barrier. An animat cannot occupy a barrier cell. The outside ring of cells, five deep, are barrier cells, locking the animats into their environment. There are 3 world types constructed from various configurations of the barriers cells (Figure 4). Open cells can contain one food unit. Food units are distributed at random throughout the open cells with a 75% probability at the beginning of each animat's life span. Food is not added to the environment during the life span of an animat. Time in the environment is measured in "clicks". In each click the state of the 13 cells of the environment in front of the animat are encoded (Figure 5a) and fed into the animat's neural network as input, the output of the net is decoded (Figure 5b) and the animat's location and orientation are updated, as well as the new state of the environment.

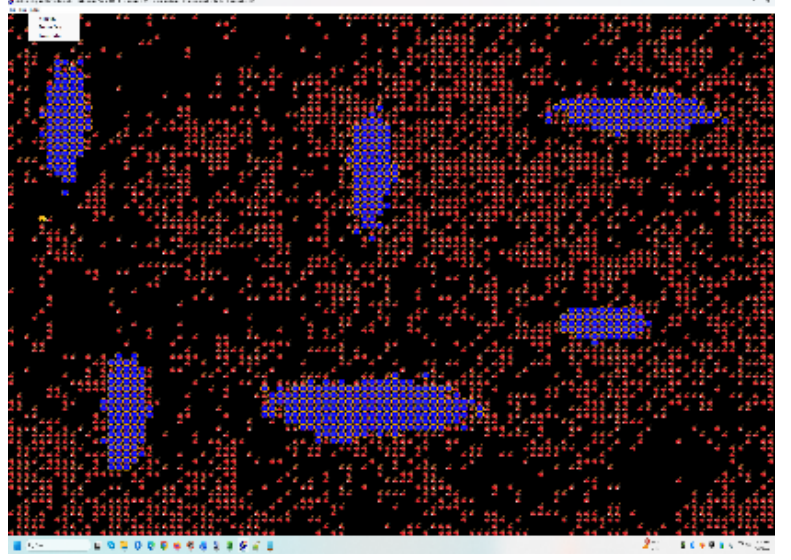
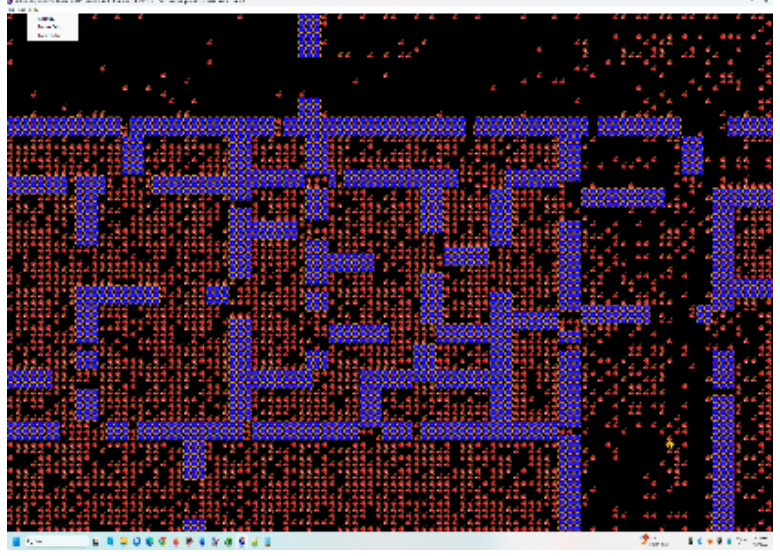
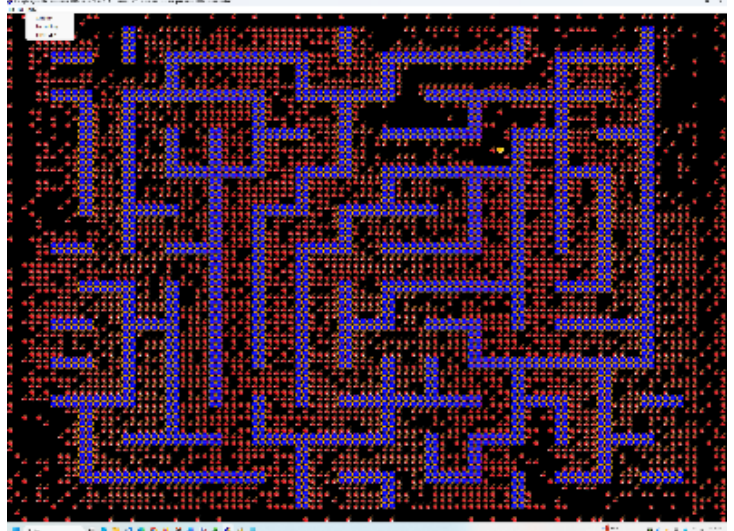

Figure 4: Artificial environments, with food and animat during an animat's life span. Blue cells are barrier, cherry's are food, animat graphic is pac-man. Delay option is to slow the animats down enough to be able to observe behavior.

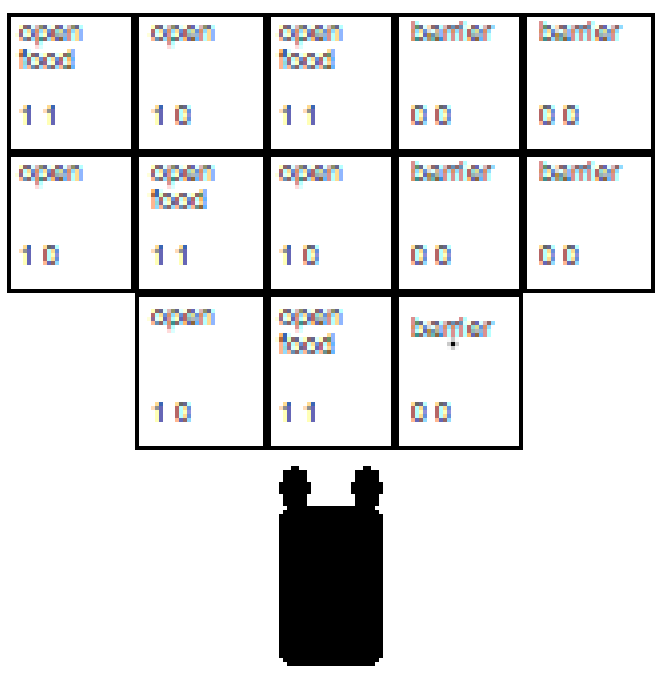

Resulting input: 10110010111000001110110000

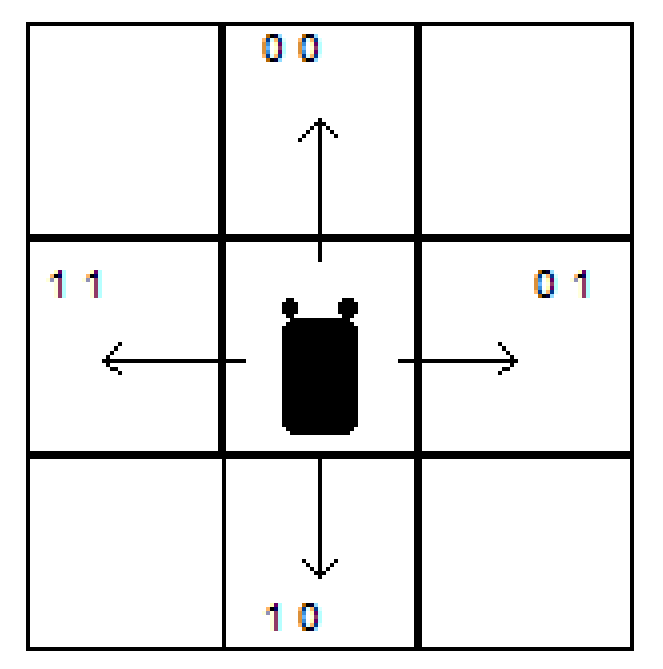

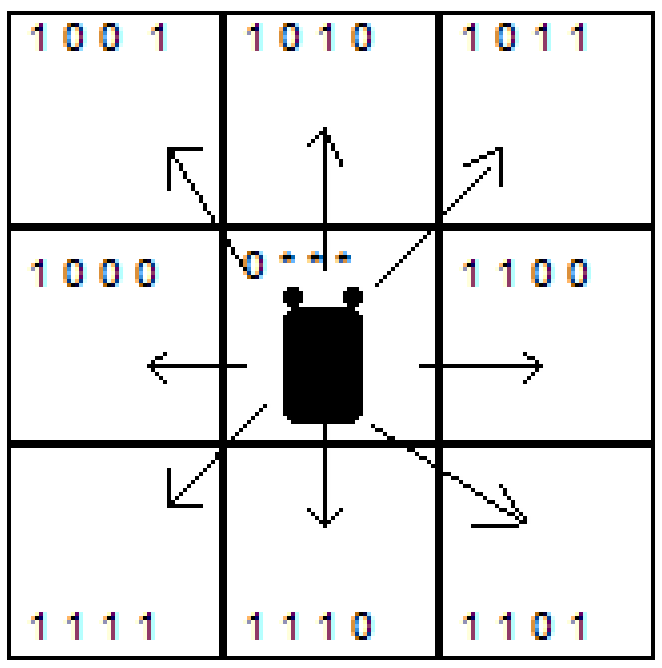

Output for turning Output for movement
Example output for moving strait ahead: 0 0 1 0 1 0

Figure 5a: Encoding of animat input data Figure 5b: Decoding of animat output data

### 3.1.2 Animats

Animats consist of an input sensor, an output register, a genotype and a hebbian neural network of 256 neurons, the topology of which is determined by the animat's genotype. At the beginning of the animat's life span the animat's genotype is read, the corresponding hebbian network is built. and the animat is placed at random in the environment. Life spans are set at 8000 clicks, although this could be lowered if the animats become sufficiently fit enough to collect all of the food in the environment before the end of their life span. Animats that do not collect any food for a continuous 256 clicks are considered to have starved to death and the animat's life span is terminated. Animats interface with their environment through the input sensor and output register. The input sensor encodes the state of the 13 cells directly in front of the animat, into a binary array, (Figure 5a). This array is given to the animat's neural network as input, and the resulting output from the neural net is put into the animat's output register. The output register is then decoded to determine the animat's intended movement and orientation, (Figure 5b). If the movement is legal, i.e. not into a barrier cell, the animat's location is updated. In any event the animat's orientation is updated. An animat collects food by moving onto a cell which contains food, after which the cell is updated to reflect that its food content is zero. An animat's fitness is calculated by the amount of food collected by the animat in its life span, higher food counts meaning higher fitness.

## 3.2 Genetics

Since part of the experiment is to test whether the method used to encode the network topology into a genotype affects the performance of the genetic algorithm, genetic operators, (selection, crossover, etc.) are kept as constant as possible across all encoding methods. Therefore any difference in performance should be due to the encoding method used. The experiment uses Matrix and Lsys encoding. The job of the genetic algorithm is to optimize the topology of the hebbian networks to improve performance on a particular problem. The animat's genotype determines the topology of its neural network. A genotype will specify the connections between the neurons as well as whether each connection is soft (learning) or hard (fixed weight) connection. If the connection is hard the genotype will also specify the weight on the connection. Genotypes are generated randomly at the beginning of a run. However the initial number of connections between the neurons, and the percentage of hard connections, is controlled.

### 3.2.1 Genetic operators

Elitism will be considered a genetic operator. Counting elitism, four genetic operators are employed in the experiment, selection, crossover, mutation, and elitism. Crossover, and elitism are performed in the standard manner, however because of the genetic alphabets chosen the mutation operator has to be customized for each encoding method. The selection mechanism is designed to discourage premature convergence and convergence stall late in the run when fitness' tend to vary less. At the end of a generation, animats are ordered by their fitness. The genotype of the fittest animat is copied into the next generation as child one. Then the genotype is mutated and copied to the next generation as child two. The same procedure is performed for the second most fit animat, with the children being three and four. After this the fittest one fourth of the population is mated four times with an animat picked at random from the fittest one third of the population. Half of these children are mutated. Mutating only half of each population reduces the probability of mutation stall.

### 3.2.2 Parameter Settings

In all genetic programs the values must be chosen for parameters such as the probability of mutation, the probability of crossover and the size of the population. These parameters can chosen by educated guess or by experimentation. The experiment is computationally expensive and therefore only limited experimentation was possible. The probability of mutation used for test runs varied from .5% to 2%. Other parameters important to this experiment include the learning coefficient for the hebbian learning rule, the connection probability, and the hard connection probability. These parameters are kept constant for Matrix vs LSys throughout all the runs. The learning coefficient is set at 0.0035, however weights are updated only during the infancy state of an animat. The infancy state consists of the first 800 clicks in the life span of an animat. The connection probability and hard connection probability are used during the initial random generation of the population at the beginning of a run. During mutation a separate tunable variable is used to control the probability of mutation in type of connection. This keeps the mutation itself from driving the number of connections and hard connections between neurons upward or downward.

### 3.2.3 Encoding method one (Matrix encoding)

Encoding method one is based upon the standard neural network connectivity matrix. In addition a second component, the weight value of the connection, is added to each cell in the matrix. The number of neurons used is 256 therefore the resulting matrix has 256 squared or 65536 cells. Each cell contains both a boolean value, zero or one, and one member of the set [-10 .. 10, 100]. The boolean values specify the connections between the neurons. A zero or 100 in the integer value specifies a soft connection. If the integer is not zero or 100 it is divided by 10 to obtain the weight on the corresponding connection.

Let n be the number of neurons in the ANN generated by the genotype. Like neural network connectivity matrices, the size of genotypes under method two grow at $O(n^2)$. Encoding method one has the advantage however of being able to specify every possible network topology.

A formal definition of Encoding method one is as follows :

**Terminology**

n = number of neurons.

C = 0 or 1.

W = integer value from the set [-10 … 10, 100], with 0 meaning a soft connection and 100 meaning adult learning soft connection. If W <> 0 or 100 then the associated connection is hard wired with weight W/10.

connection pair = C W, which is expressed as a neural network connectivity matrix.

### 3.2.4 Encoding method two (LSys GNN encoding method)

Genotypes in encoding method two are sets of L-System productions. Multiple sets of productions per genotype can be used. Wp1hgn uses two productions. The first generates a standard neural network connectivity matrix when expanded. The second generates the weight component. Together they are expanded to form a connectivity/weight matrix similar to the one used in encoding method one. In order to keep the number of neurons constant a constrained L-System model is used. L-symbols are classified by level, with each L-symbol expanding to exactly four L-symbols from the level beneath it. This results in expansions of constant size. Both sets of productions use the same symbols until layer $\log_2(n)$-1, which contains the terminal symbols. The $\log_2(n)$-1 layer in both sets of rules consists of sixteen L-symbols, the set [a..p]. In the first set of rules the sixteen symbols are hard coded to expand to the sixteen possible binary strings of length four, this is the $\log_2(n)$ connectivity layer. In the second set the $\log_2(n)$ layer consists of numerical weight values. Genotypes consist of multiple strings of L-symbols from each level. Each string is the right side of an L-System production. The left side of the L-System productions are implicit in the ordering of the L-symbols. The right side of layer $\log_2(n)$-1 is hard coded in the first set of rules and thus is not included in the genotypes.

Using Lsys encoding the growth in the size of the genotypes in relation to the number of neurons, n, is $O(\log_2 n)$.

**A formal definition of the Lsys GNN encoding method :**

Terminology

$X_i$ = L-symbol variable of level i.
n = number of neurons.
W = independently specified integer weight value.

L-symbols

$A_1,B_1,C_1,D_1$ = level one L-symbol.
$A_2,B_2,C_2,D_2$ = level two L-symbol.
$A_3,B_3,C_3,D_3$ = level three L-symbol.
$A_4,B_4,C_4,D_4$ = level four L-symbol.
....
$A_i,B_i,C_i,D_i$ = level i L-symbol, where $i=\log_2(n)-2$.
a,b,c,...,p = terminal level L-symbol, where $i=\log_2(n)-1$.

Each L-symbol from level i is expanded to a string of four L-symbols from level i+1.

- Ai -> Level[i+1][1..4]
- Bi -> Level[i+1][5..8]
- Ci -> Level[i+1][9..12]
- Di -> Level[i+1][13..16]

For the final expansion, level $\log_2(n)$:

If the expansion rule set is a connectivity matrix then

a is expanded to 0 0 0 0
b is expanded to 0 0 0 1
c is expanded to 0 0 1 0
....
p is expanded to 1 1 1 1

If the expansion rule set is a weight matrix then

a is expanded to W W W W [1..4]
b is expanded to W W W W [5..8]
c is expanded to W W W W [9..12]
....
p is expanded to W W W W [61..64]

An example Lsys encoding genotype for 256 neurons:

```
[Herbivore 1]
CABC
CDADBACCDBCADDCD
CBAABCCAAADCAADD
ABABABBBBCABCBCC
CDBBCBAACABBDDDB
CCBCCDBBBCADBDAB
dojfedepjdanfmnp
```

```
DDBD
CDDABABACCABACAA
CCDBDCBDBAAACBAA
AACBBDCCCCABAABC
BCBDAABCBDDDBCBB
CCAAAACADADBAAAA
dnndddnnbbddindd
0 10 0 0 -4 5 -10 -5 100 -2 100 -4 -5 -10 -10 -10 0 0 100 0 100 100 -2 0 0 0 0 0 0 0 100
100 0 4 0 -10 100 -1 0 0 0 0 0 0 0 0 100 -8 100 -8 0 100 -5 -10 -4 -10 0 0 -3 0 0 0 2 100
```

All symbols in layers 1 - $\log_2(n)$-2 can be comprised of the same four distinct characters.  Location specifies level.

Wp1hgn use case of W: W = integer value from the set [-10 … 10, 100], 0 = soft connection, 100 = adult learning soft connection.  If W <> 0 or 100 then the associated connection is set to hard connection with weight W/10.

### 3.2.5 Wp1hgn

To run the experiment a neural network genetic modeling tool titled Wp1hgn was designed and built.  Wp1hgn can generate populations of random genotypes of the Matrix and Lsys alphabets with varying parameters and and then perform genetic optimization on multiple world types.   Options for mutation rate, number of crossover points during combination and other parameters can be chosen before or during each run.  Wp1hgn will be put up on https://www.bio.fsu.edu/astuy/Wp1hgn and FigShare, allowing other researchers to test the results.

### 3.2.6 MatrixLSG

To test whether any increase in performance for Matrix vs Lsys populations is due to the genetic alphabet or the initial generation of the random genotypes, an option for generating the initial random Matrix genotypes by first generating random Lsys genotypes and then converting them to Matrix genotypes is available in Wp1hgn.

## 3.3 Results

Fitness of the animats (F) is measured by the amount of food units collected in the animats life span. Performance of each run is measured by both the maximum and average amount of food collected by the animats in a generation, Max($F_i$), 1 <= i <= popsize   ($\sum F_i$) / popsize, 1 <= i <= popsize, as well as the number of generations required to achieve the result.  At the beginning of a run performance is very poor, as would be expected.  Most animats perform tight circles, or crash into barriers and stick.  Behavior rapidly improves in the first generations.  The animats learn to move forward and avoid barriers.  Actively seeking food is usually the last behavior to be learned.   Lsys encoding is shown to provide faster and more consistent learning as well as more consistent and adaptable behavior.

All 8 Lsys populations achieved maximum food collection performance more than twice that of random movement animats by generation 1000, with a mean maximum food count of 3802 ± 197 (n=8, CV=5.2%). Every Lsys run first exceeded 2000 food collected between generations 13 and 33, demonstrating rapid and consistent convergence. In contrast, only 4 of 8 Matrix runs ever exceeded 2000 food collected at any point during the 1000 generation run, with the earliest occurring at generation 151 and the latest at generation 667. The remaining 4 Matrix runs never exceeded 2000 food collected, representing complete failure to achieve competitive performance. (Figures 6, 7).

While Matrix encoding occasionally produced competitive animats, performance was highly inconsistent across runs. At generation 1000, Matrix encoding achieved a mean maximum food count of 1388 ± 610 (n=8, CV=44.0%), compared to 3802 ± 197 for Lsys encoding — a 2.74x advantage in mean performance. The coefficient of variation for Matrix encoding (44.0%) was 8.5 times that of Lsys encoding (5.2%), quantifying the dramatic difference in reliability between the two encoding methods. Population average food count at generation 1000 showed a similar pattern: Lsys achieved 3207 ± 181 versus Matrix at 513 ± 252, a 6.3x advantage in mean population-wide performance. (Figures 6, 7).

To test the generalization ability of evolved populations, the top 4 populations of each encoding type were transferred to Maze world, a novel environment with narrow corridors and dead ends presenting substantially different navigational challenges than the RoundedBarrier1 training environment. Each population was run for 100 generations in the new environment. Lsys populations demonstrated immediate and robust transfer, entering Maze world at generation 1 already collecting a mean maximum food count of 2447 and maintaining strong performance throughout, finishing at 2455 ± 176 at generation 100. In striking contrast, Matrix populations collapsed in the novel environment, achieving a mean maximum food count of only 422 ± 212 at generation 100, a 5.82x disadvantage compared to Lsys populations and substantially worse than their already poor training world performance. MatrixLSG populations showed intermediate transfer performance of 1274 ± 506 at generation 100. (Figure 8.) The Lsys advantage over Matrix encoding increased from 2.74x in the training environment to 5.82x in the novel environment, demonstrating that Lsys encoding produces not merely better optimized but fundamentally more transferable and robust neural network behaviors.

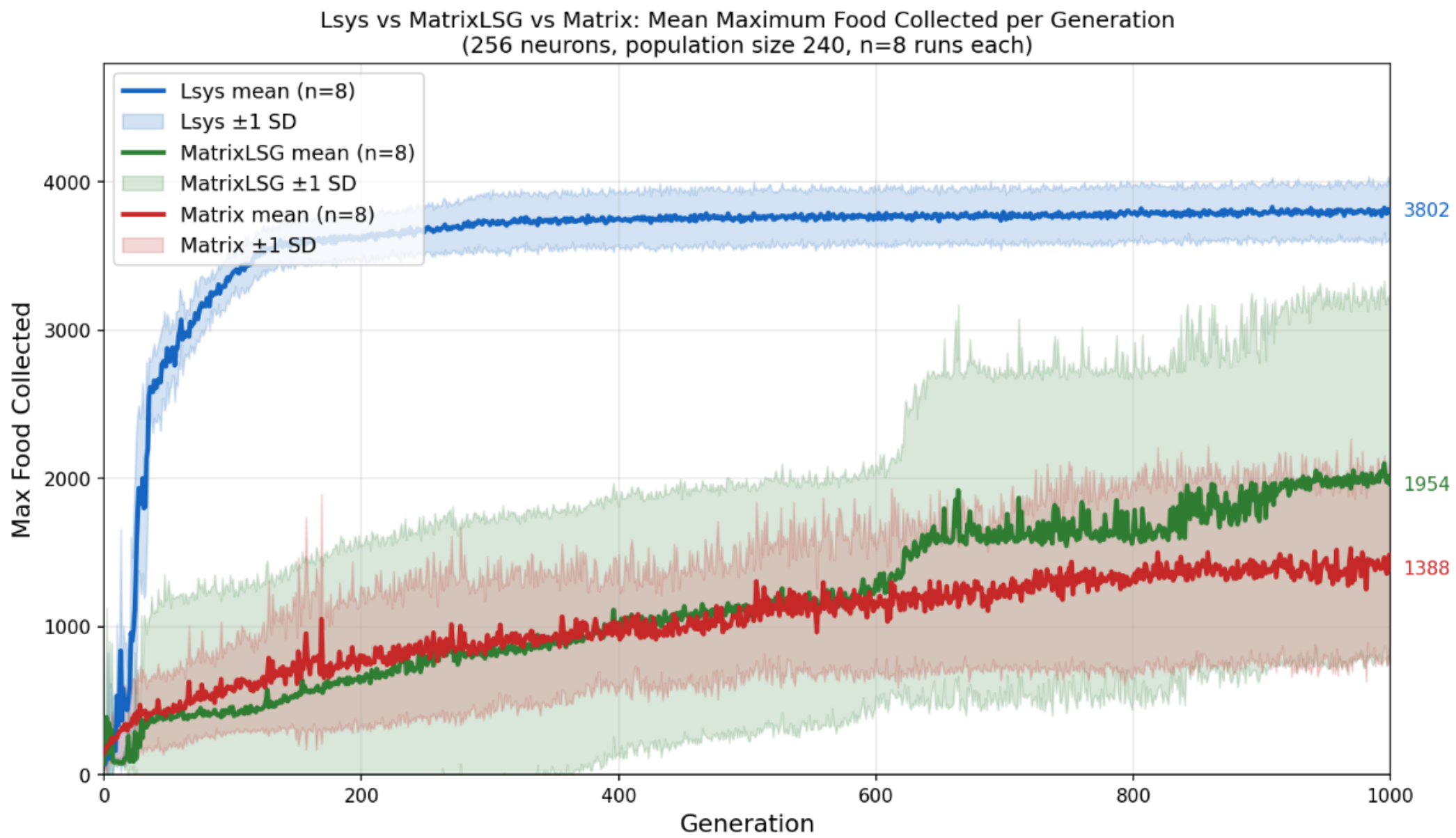

Figure 6. Lsys vs MatrixLSG vs Matrix: Mean Maximum Food Collected Per Generation

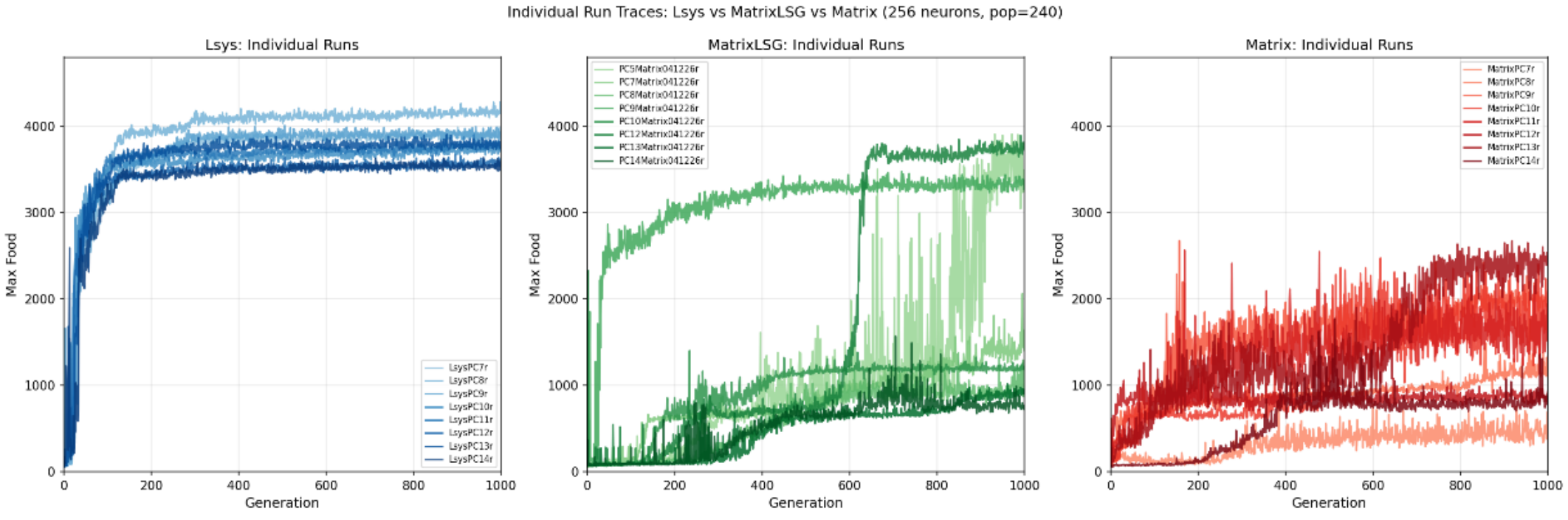

Figure 7. Individual Run Traces: Lsys vs MatrixLSG vs Matrix

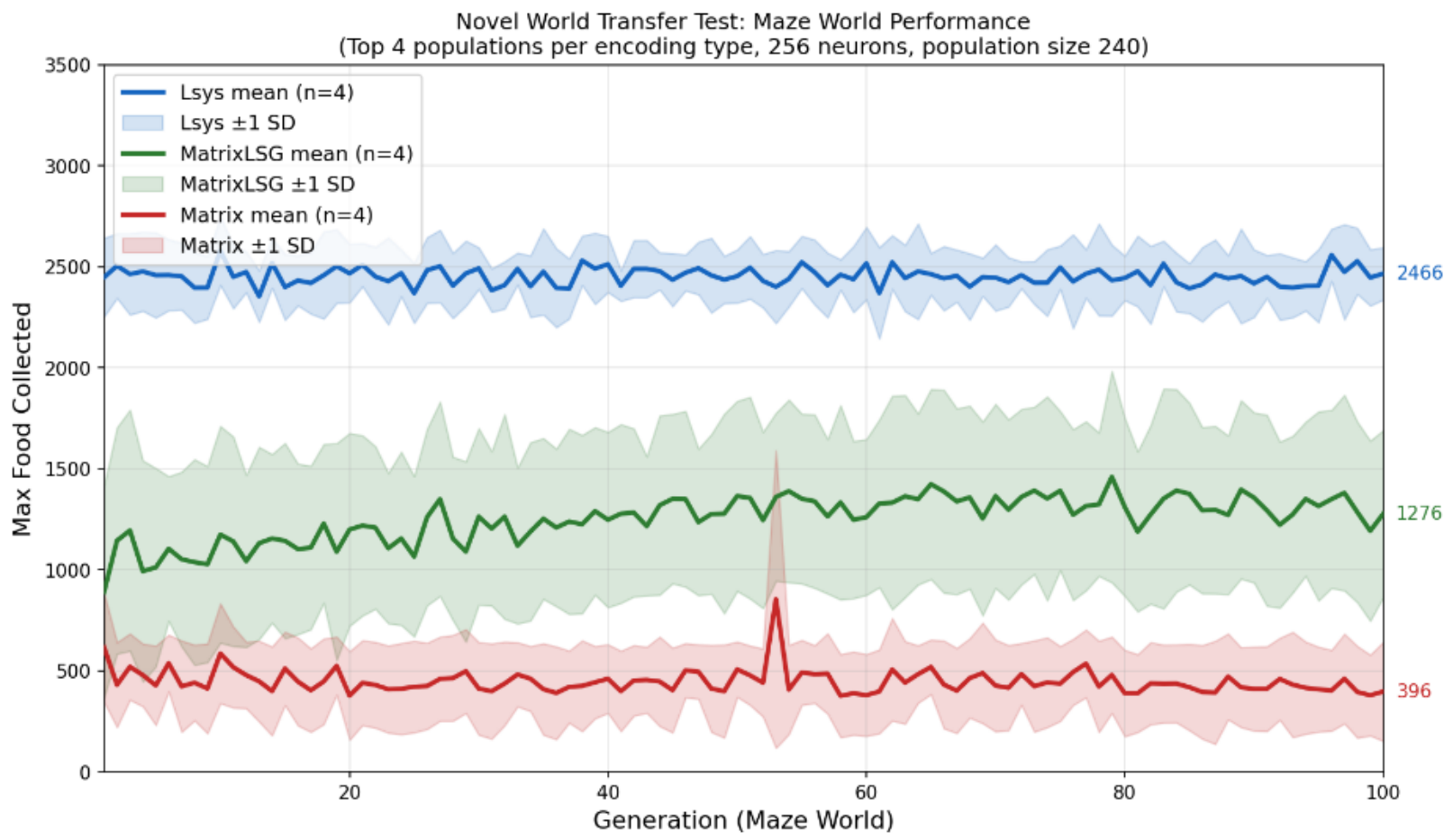

Figure 8. Novel World Transfer Test

# 4 Summary and Conclusions

## 4.1 Summary

A formal L-System based genetic alphabet for neural networks, titled Lsys, that scales at $\log_2(n)$ was introduced. A neural network genetic modeling tool, Wp1hgn, was introduced as a method for optimizing hebbian neural networks using genetic algorithms and testing encoding methods for the neural network topologies. An artificial environment was used to test network performance. All runs of the experiment resulted in increased network performance in the artificial environment after sufficient generations. By combining hebbian ANNs and GAs hebbian ANNs were produced that could successfully navigate an artificial environment while searching for food. The ANNs were never given teaching data, they evolved an instinctive solution to fitness in their environment.

Two methods for encoding hebbian network topologies were tested, Lsys and Matrix. Lsys encoding achieved a mean maximum food count of 3802 ± 197 at generation 1000 compared to 1388 ± 610 for Matrix encoding, a 2.74x performance advantage with an 8.5-fold improvement in consistency. Genotypes generated via Lsys and then converted to Matrix format (MatrixLSG) achieved a mean maximum food count of 1954 ± 1178 (CV=60.3%), a modest 1.41x improvement over direct Matrix encoding that falls far short of the 2.74x advantage achieved by full Lsys encoding, showing that the increase in learning for Lsys genotypes is due to the genetic algorithm operating on the Lsys genetic alphabet, not the initial generation of the genotypes. Lsys encoding also performed dramatically better on novel world types, achieving a mean maximum food count of 2455 ± 176 in Maze world compared to 422 ± 212 for Matrix encoding, a 5.82x advantage that exceeded the 2.74x advantage observed in the training environment, demonstrating that Lsys encoding produces fundamentally more transferable neural network behaviors.

## 4.2 Conclusions

The experiment demonstrates that genetic algorithms can successfully optimize hebbian neural networks for navigation in an artificial environment without any supplied training data, and that the choice of genetic encoding method has a profound effect on both the performance and reliability of the optimization process.

Lsys encoding proved decisively superior to Matrix encoding across all measured dimensions. Lsys populations achieved a mean maximum food count of 3802 ± 197 at generation 1000, compared to 1388 ± 610 for Matrix encoding — a 2.74x performance advantage. More significantly, every Lsys population successfully learned to navigate the environment, while 4 of 8 Matrix populations failed to achieve competitive performance at any point during 1000 generations. The consistency advantage is significant: Lsys encoding produced a coefficient of variation of 5.2% across 8 runs with varied parameters, compared to 44.0% for Matrix encoding, 8.5 times more variable. This robustness to parameter variation suggests that the Lsys genetic alphabet provides structural guidance to the evolutionary search that is largely independent of specific parameter settings.

The MatrixLSG control condition isolates the source of the Lsys advantage. Populations initialized from Lsys-generated networks but evolved with Matrix operators achieved a mean maximum food count of 1954 ± 1178 at generation 1000 (CV=60.3%), only a modest 1.41x improvement over direct Matrix encoding and far below the 2.74x advantage of full Lsys encoding. Notably the MatrixLSG coefficient of variation (60.3%) was actually higher than that of Matrix encoding (44.0%), confirming that Lsys-structured initialization alone does not provide the consistency advantage seen in full Lsys runs. These results demonstrate that the performance advantage of Lsys encoding derives primarily from the genetic algorithm operating continuously on the compressed symbolic Lsys alphabet throughout evolution, rather than from the structure of the initial population.

The novel world transfer test further confirms the superiority of Lsys encoding. When transferred to Maze world, a novel environment with narrow corridors and dead ends, Lsys populations immediately demonstrated robust navigational competence with no adaptation period, while Matrix populations collapsed to near-random performance. The Lsys advantage increased from 2.74x in the training environment to 5.82x in the novel environment, suggesting that Lsys encoding produces genuinely general navigational strategies rather than environment-specific optimizations. MatrixLSG populations again showed intermediate performance (1274 ± 506), with the stronger training world performers showing meaningful transfer while weaker ones failed, further supporting the conclusion that it is the ongoing operation of genetic operators on the symbolic Lsys alphabet, rather than initialization structure alone, that produces robust and transferable learned behaviors.

These results confirm the theoretical prediction that $O(\log_2 n)$ symbolic encoding methods can provide superior performance over $O(n^2)$ direct encoding methods for neuroevolution. The Lsys alphabet introduced in this paper provides a practical and highly effective implementation of this principle for hebbian neural networks, achieving faster convergence, higher peak performance, dramatically greater reliability, and superior generalization to novel environments compared to matrix encoding across all experimental conditions tested.

### 4.3 Areas for further research

Short term: The Lsys genetic alphabet scales at $O(\log_2 n)$. Future versions of Wp1hgn will test whether learning for Lsys networks scales at $O(\log_2 n)$ for larger neural network sizes. Wp1hgn currently operates on neural neworks of 256 neurons. Capability will be added for Wp1hgn to support neural networks of 512, 1024, 2048 and 4096 neurons using the Lsys alphabet. The Lsys alphabet currently supports a very limited variety of L-System expansion rules, limiting the expression of possible network topologies. Research into semantic expansion of the Lsys alphabet may result in further increases in Lsys performance.

Long term: While this experiment uses Hebbian neural networks, L-System encoding of other types of neural networks, including supervised learning models is possible. The current assumption in deep learning is that more parameters plus more data equals better performance. This research raises a different question, what if a more structured compressed encoding could achieve similar or better performance with fewer parameters?

### 4.4 Data Availability Statement

All experimental data files used in this study, including the complete generation-by-generation performance records for all 24 training world runs (8 Lsys, 8 Matrix, and 8 MatrixLSG) and all 12 novel world transfer test runs, are publicly available at https://www.bio.fsu.edu/astuy/Wp1hgn/data and on FigShare at https://figshare.com/authors/Alexander_Stuy/23749278. The Wp1hgn simulation software, a Windows 64-bit executable with installer, will be made available at https://www.bio.fsu.edu/astuy/Wp1hgn and on FigShare allowing other researchers to reproduce and extend these results.